\documentclass[letterpaper, 10 pt, conference]{ieeeconf}  

\IEEEoverridecommandlockouts                              

\usepackage{algorithm}
\usepackage{algpseudocode}
\usepackage{amsmath}
\usepackage{amssymb}

\usepackage{booktabs}      
\usepackage{multirow}      
\usepackage{graphicx}      
\usepackage{dblfloatfix}
\usepackage{xcolor}

\title{\LARGE \bf
Occluding the Solution Space: Planner-Agnostic Adversarial Attacks on Tolerance-Aware Manipulation}

\author{
Keke Tang$^{1,*,\dagger}$,   Tianyu Hao$^{1,*}$, Weilong Peng$^{1,\dagger}$, Hao Jiang$^{2}$,\\
Feng Wu$^{2}$, Peican Zhu$^{3,\dagger}$, Jianmin Ji$^{2}$, and Zhihong Tian$^{1}$%
\thanks{$^{*}$These authors contributed equally. $^{\dagger}$Corresponding authors.}%
\thanks{$^{1}$Keke Tang, Tianyu Hao, Weilong Peng, and Zhihong Tian are with Guangzhou University, Guangzhou 510006, China.
$^{2}$Hao Jiang, Feng Wu, and Jianmin Ji are with the University of Science and Technology of China, Hefei 230026, China.
$^{3}$Peican Zhu is with Northwestern Polytechnical University, Xi'an 710072, China.
}%
\thanks{This work was supported in part by the National Natural Science Foundation of China (62472117, 62572400, U2436208, 62372129), 
the Guangdong Basic and Applied Basic Research Foundation (2025A1515010157, 2024A1515012064), 
the CCF-NetEase ThunderFire Innovation Research Funding (CCF-Netease 202514),
the Science and Technology Projects in Guangzhou (2025A03J0137, 2024B0101010002).}%
}

\begin{document}

\maketitle
\thispagestyle{empty}
\pagestyle{empty}

\begin{abstract}
Adversarial attacks on motion planning are crucial for evaluating and quantifying the intrinsic robustness of robotic manipulation. However, existing approaches are typically limited by  restrictive exact-pose objectives and their reliance on  planner-in-the-loop queries.
To address these limitations, we propose a planner-agnostic attack framework for tolerance-aware manipulation. Our approach shifts the evaluation paradigm to task-level feasibility over goal regions, efficiently inserting adversarial obstacles without requiring oracle access to the victim system. Offline, we characterize the robot's intrinsic workspace capabilities via a kinematic occupancy heatmap, which encodes the density of feasible trajectories and robustness priors without invoking a specific planner. Online, we formulate the attack as a budgeted maximum-coverage optimization, strategically deploying obstacles subject to explicit geometric constraints to occlude the solution space. Extensive experiments across simulation and real-world scenarios demonstrate that our method reliably induces planning failures, significantly outperforming planner-in-the-loop baselines in both computational efficiency and attack efficacy. 
\end{abstract}

\section{Introduction}
\label{sec:intro}

Robotic manipulation relies fundamentally on motion planning to translate high-level task specifications into safe and feasible arm trajectories~\cite{kuffner2000rrtconnect,karaman2011sampling,9669009}.
In safety-critical deployments, the reliability of this planning process is non-negotiable, as failures can lead to operational hazards or costly interruptions.
Adversarial attacks, which utilize targeted environmental modifications to provoke system failures, provide a rigorous framework for probing these safety boundaries.
By systematically generating worst-case scene configurations, such methods serve as a necessary stress test to quantify the intrinsic robustness of planning algorithms.

Despite their importance, adversarial evaluations targeting classical manipulation planners remain underexplored.
Prior investigations have predominantly focused on perception-induced failures~\cite{sadeghi2023attacking}.
The limited body of work addressing planning robustness typically enforces highly restrictive success criteria, such as tracking specific pose targets within tight tolerances~\cite{wu2024characterizing}.
However, these \emph{exact-pose objectives} confound task-level feasibility: a plan functionally fails only if it misses the grasp region entirely, not merely a specific coordinate.
Furthermore, existing approaches commonly rely on \emph{planner-in-the-loop} optimization~\cite{wu2024characterizing}.
This dependency grants the adversary unrealistic oracle-like access to the victim planner, rendering the threat model impractical for black-box security assessments.

\begin{figure}
    \centering
    \includegraphics[width=0.9\linewidth]{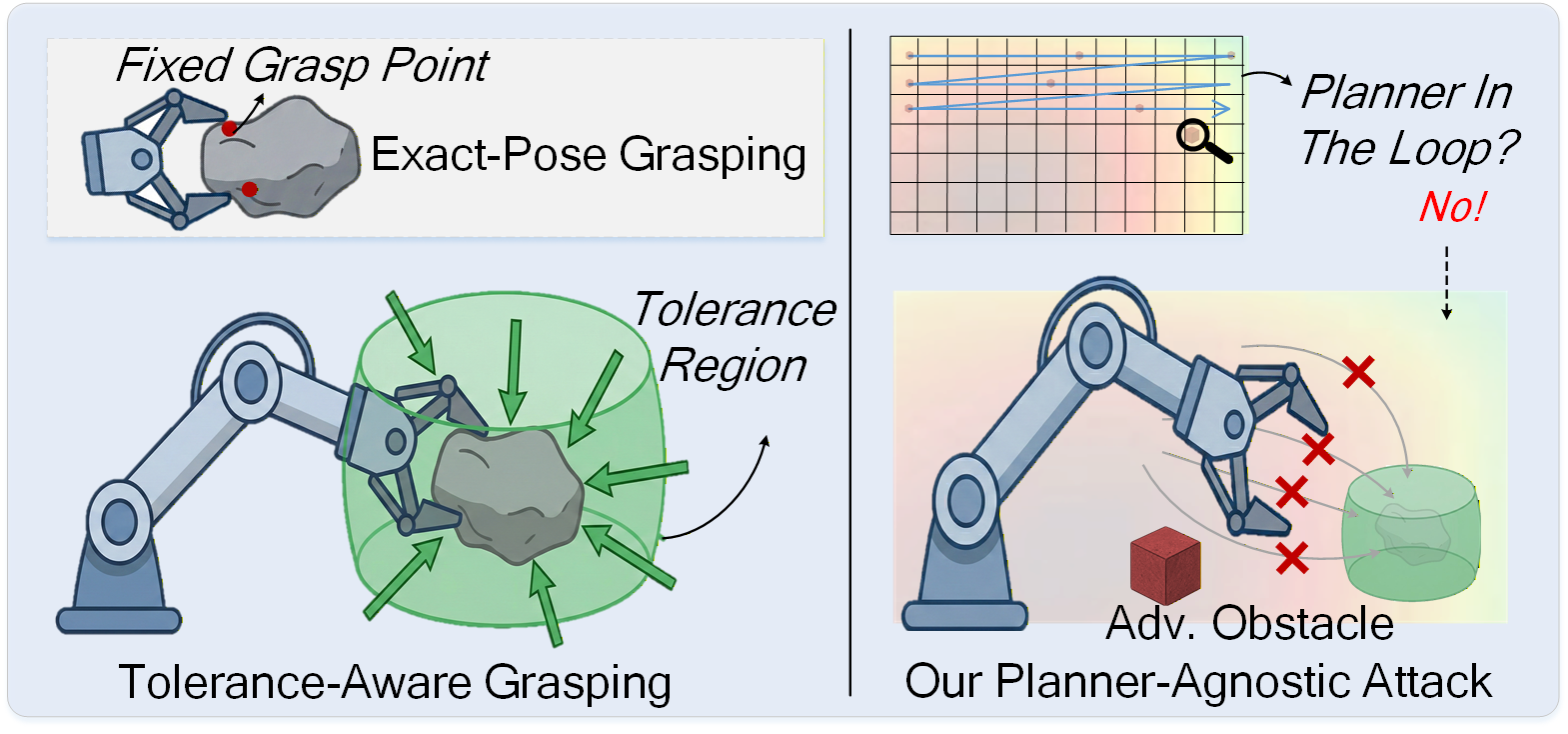}
    \vspace{-3mm}
    \caption{Concept of the proposed planner-agnostic attack on tolerance-aware manipulation. Unlike conventional exact-pose grasping that targets fixed points (top-left), we consider a more realistic tolerance-aware task where the goal is a continuous region (bottom-left). Our framework (right) generates adversarial obstacles to occlude the robot's potential motion paths without requiring oracle access to the victim's internal planner.}
    \label{fig:teaser}
\end{figure}

To address these limitations, we argue that a practical adversarial framework must be fundamentally rethought across two critical dimensions: the task objective and the threat model.
From a task perspective, manipulation is inherently \emph{tolerance-aware}: operational success depends on reaching a feasible goal region, allowing the planner to utilize a continuous manifold of valid configurations, see Fig.~\ref{fig:teaser}.
Consequently, an effective attack must occlude this entire solution space rather than merely obstructing a singular path.
From a threat model perspective, a realistic adversary must be \emph{planner-agnostic}, operating without oracle access to the victim system.
Instead, we consider an attacker that leverages only public robot kinematics and static scene observability to deploy a limited budget of obstacles subject to explicit geometric constraints, see Fig.~\ref{fig:teaser}.

In this paper, we propose a two-stage, planner-agnostic framework to execute adversarial attacks on manipulation tasks by systematically occluding the feasible solution space.
In the offline stage, we characterize the robot's intrinsic capabilities by constructing a \emph{kinematic occupancy heatmap}.
This volumetric prior encodes the density of valid manipulation trajectories, identifying critical spatial bottlenecks without invoking any specific planner.
During the online stage, we formulate the attack as a \emph{budgeted maximum-coverage optimization} within a target-aligned search region.
By strategically deploying obstacles, our method maximizes the occlusion of the task-level solution manifold.
This approach efficiently yields robust adversarial constraints, bypassing the computational burden of iterative planner queries while ensuring rigorous geometric validity.
Extensive experiments across diverse simulation environments and real-world robotic deployments validate the effectiveness and efficiency of the proposed method.

Overall, our  contributions are summarized as follows:
\begin{itemize}
    \item We formulate a {task-level adversarial threat model} for tolerance-aware manipulation, shifting the attack objective from targeting exact-pose configurations to occluding the entire feasible goal region.

    \item We propose a two-stage, planner-agnostic framework that efficiently generates geometric constraints by combining an offline {kinematic occupancy heatmap} with an online {budgeted maximum-coverage optimization}.

    \item We extensively validate our approach across diverse simulation environments and real-world robotic deployments, demonstrating superior attack efficacy and  efficiency compared to state-of-the-art methods.
\end{itemize}

\section{Related Work}
\label{sec:related}

\subsection{Adversarial Attacks on Robotic Systems}
\label{sec:related:robotics}

Adversarial attacks have been extensively studied in computer vision and successfully extended to the robotics domain, particularly in 3D perception~\cite{Xiang-2019-3DAdversarialPCD,Tang-2023-ManifoldAttack,tang2024symattack,tang2022rethinking,tang2026less}.
In the context of robotic manipulation, existing attacks predominantly target the grasp detection and evaluation stages.
For instance, prior work~\cite{alharthi2024physical} misleads image-based grasp evaluation networks via pixel-level perturbations or barely visible patches.
Alternatively, other approaches~\cite{wang2019adversarial,wang2025advgrasp} tackle the problem by altering the target object's physical geometry to diminish its overall graspability.

Crucially, these paradigms aim to invalidate the grasp target itself.
In contrast, our work addresses the often-overlooked vulnerability in the motion planning phase, where adversarial environmental modifications can obstruct the physical path necessary to approach and execute the task.

\subsection{Adversarial Attacks on Manipulation Planning}
\label{sec:related:planning}




Adversarial attacks on manipulation planning remain underexplored. Existing work induces failures by corrupting state estimation~\cite{sadeghi2023attacking} or exploiting semantic vulnerabilities in high-level VLM-based agents~\cite{wangadvedm, wang2025exploring} and VLA models~\cite{11560938}. These methods perturb inputs rather than evaluate planner resilience to adversarially structured environments.

Directly attacking planning algorithms via environmental constraints is a sparse field. 
To our knowledge, the only closely related work is by Wu \emph{et al.}~\cite{wu2024characterizing}, which not only requires a planner-in-the-loop approach but also relies on rigid exact-pose objectives that overlook task-level tolerance. 
In contrast, our planner-agnostic framework addresses this by efficiently generating obstacles to occlude the solution space without requiring oracle access.

\section{Problem Formulation}
\label{sec:problem}

\subsection{Tolerance-Aware Manipulation Planning}
\label{sec:problem:task}

We consider the problem of motion planning for manipulation, where the objective is to reach a \textbf{goal region} rather than a singular pose.
Let $\mathcal{Q} \subset \mathbb{R}^n$ denote the configuration space of an $n$-DoF manipulator, and let $E$ denote the nominal static 3D scene geometry.
The tolerance-aware goal is specified by a spatial subset $\mathcal{G} \subset SE(3)$, representing the volume of acceptable end-effector poses.

This task constraint induces a \textbf{feasible goal set} in the joint space, denoted as $\mathcal{Q}_{\mathcal{G}} = \{q \in \mathcal{Q} \mid \mathrm{FK}(q) \in \mathcal{G}\}$.
The planning task is to find a collision-free path starting from a fixed $q_{\mathrm{init}}$ and terminating at any configuration $q_{\mathrm{end}}$ that satisfies:
\begin{equation}
q_{\mathrm{end}} \in \mathcal{Q}_{\mathcal{G}}.
\label{eq:goal_region_pf}
\end{equation}
Crucially, unlike classical point-to-point planning where the goal is a singleton, $\mathcal{Q}_{\mathcal{G}}$ forms a continuous manifold. This redundancy provides the planner with multiple valid solutions, making the task inherently more resilient to simple constraints.

\subsection{Adversarial Threat Model}
\label{sec:problem:threat}

The objective of the adversary is to \textbf{force a task failure} by strategically modifying the environment. Given the redundancy defined in Sec.~\ref{sec:problem:task}, a successful attack must effectively \textbf{occlude the entire solution space} that connects $q_{\mathrm{init}}$ to the manifold $\mathcal{Q}_{\mathcal{G}}$. We formulate this as a constrained geometric optimization problem.

\paragraph{Attack Action}
The adversary generates a set of $M$ obstacles $\mathcal{O}(\theta) = \{o_1(\theta_1), \ldots, o_M(\theta_M)\}$ parameterized by $\theta$, resulting in a perturbed environment:
\begin{equation}
E_\theta := E \cup \mathcal{O}(\theta).
\label{eq:env_theta_set}
\end{equation}
The attack is successful if, for a given environment $E_\theta$, no path $\tau$ satisfies the conditions in Eq.~\eqref{eq:goal_region_pf}.

\paragraph{Insertion Constraints}
To ensure the attack remains physically realistic and maintains scene consistency, the parameters $\theta$ are subject to the following constraints:
\begin{itemize}\setlength\itemsep{0.1em}
    \item[\textbf{C1)}] \textbf{Cardinality Budget:} The adversary is restricted to inserting at most $M$ distinct obstacles.
    \item[\textbf{C2)}] \textbf{Geometric Bounds:} Each obstacle $o_m$ is defined by standard geometric primitives (e.g., boxes or spheres) with a maximum  size.
    \item[\textbf{C3)}] \textbf{Placement Feasibility:} Insertions must lie within a valid search region $\mathcal{S} \subset E_{\mathrm{free}}$ and maintain a minimum clearance from $q_{\mathrm{init}}$ and $\mathcal{G}$ to avoid trivial infeasibility.
\end{itemize}

\begin{figure*}[!t]
    \centering
    \includegraphics[width=0.99\linewidth]{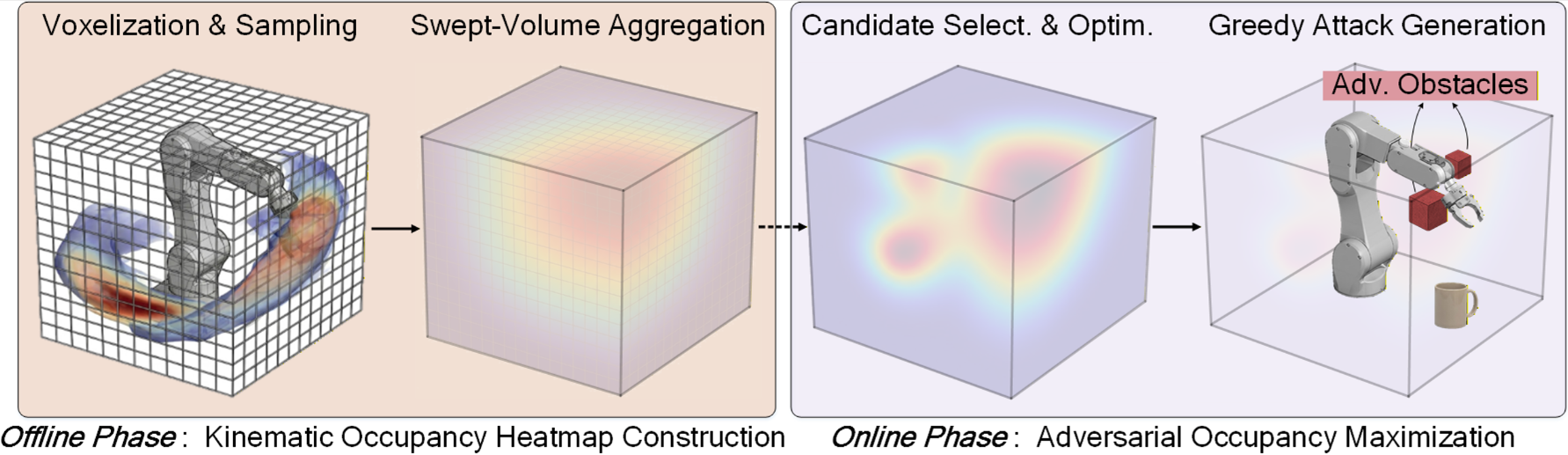}
    \caption{Framework of the planner-agnostic attack. The offline phase constructs a kinematic occupancy heatmap to characterize workspace importance. The online phase then leverages this heatmap to perform budgeted maximum-coverage optimization for generating adversarial obstacles (red boxes) that disrupt tolerance-aware manipulation planning.}
    \label{fig:framework}
\end{figure*}

\paragraph{Adversary Capabilities}
We assume a {planner-agnostic} adversary operating under the following conditions:
\begin{itemize}\setlength\itemsep{0.1em}
    \item[\textbf{A1)}] \textbf{Known Kinematics:} The adversary has access to the robot's kinematic model and collision geometry.
    \item[\textbf{A2)}] \textbf{Scene Observability:} The adversary perceives the nominal scene $E$ and the task specification $\mathcal{G}$.
    \item[\textbf{A3)}] \textbf{No Oracle Access:} The adversary {cannot} query the victim planner or simulate the full planning process during attack generation. This ensures the attack exploits intrinsic geometric vulnerabilities rather than over-fitting to a specific algorithm's internal logic.
\end{itemize}


\section{Methodology}
\label{sec:method}

We propose a \textbf{planner-agnostic framework} to generate adversarial constraints that occlude the workspace volume most critical to the solution space of tolerance-aware manipulation. 
Our approach operates in two stages: 
(i) an \textbf{offline phase} that constructs a kinematic occupancy heatmap characterizing the robot's intrinsic capabilities; 
(ii) an \textbf{online phase} that formulates the attack as a budgeted maximum-coverage optimization to strategically place obstacles.
Please refer to Fig.~\ref{fig:framework} for an illustration.

\subsection{Offline: Kinematic Occupancy Heatmap Construction}
\label{sec:method:offline}

The offline stage maps the high-dimensional configuration space $\mathcal{Q}$ into a workspace scalar field $\mathcal{H}$. This field identifies ``spatial bottlenecks''---regions where valid and robust arm configurations are most densely concentrated.

\subsubsection{Workspace Voxelization}
The workspace is discretized into a voxel grid $\mathcal{V}$ with resolution $\delta{\times}\delta{\times}\delta$. Let $\mathcal{V}_{\mathrm{free}} \subset E_{\mathrm{free}}$ denote the set of voxels not occupied by static geometry $E$. We initialize a heatmap $\mathcal{H}: \mathcal{V}_{\mathrm{free}} \rightarrow \mathbb{R}_{\ge 0}$ to zero, which will serve as the importance measure for obstacle placement.

\subsubsection{Robust Configuration Sampling}
We sample a large set of configurations $q \in \mathcal{Q}$ and retain only those that satisfy self-collision and scene-collision constraints: $\mathrm{valid}(q; E) = \mathrm{True}$. 
To ensure the heatmap reflects regions favored by robust motion planners, each valid sample is assigned a weight $w(q)$ combining two quality metrics:
\begin{equation}
w(q) = s_{\mathrm{man}}(q)^{\gamma_1} \cdot s_{\mathrm{clr}}(q)^{\gamma_2},
\end{equation}
where $\gamma_1, \gamma_2 \ge 0$ are hyperparameters. 

\paragraph{Manipulability Score ($s_{\mathrm{man}}$)}
We utilize the Yoshikawa index, $m(q) = \sqrt{\det(J(q)J(q)^\top)}$, where $J(q)$ is the geometric Jacobian. We apply a saturation function to bound the score: $s_{\mathrm{man}}(q) = m(q) / (m(q) + \kappa)$. 
The purpose of this score is to identify regions where the robot possesses high control authority. By prioritizing these regions, the adversary targets configurations furthest from kinematic singularities, which motion planners preferentially utilize to maintain trajectory responsiveness and maneuverability.

\paragraph{Clearance-Robustness Score ($s_{\mathrm{clr}}$)}
To prioritize configurations in open workspace areas, we compute the minimum distance between the robot's occupied volume $\mathcal{B}(q)$ and static obstacles $E$, denoted $d_{\mathrm{clr}}(q) = \mathrm{dist}(\mathcal{B}(q), E)$. This is normalized using a safety margin $d_{\mathrm{safe}}$:
\begin{equation}
s_{\mathrm{clr}}(q) = \min\!\left( d_{\mathrm{clr}}(q)/d_{\mathrm{safe}}, 1 \right).
\end{equation}
By occluding these high-clearance regions, which planners naturally favor, the adversary effectively forces the robot into narrow, constrained corridors where planning is  harder.

\subsubsection{Swept-Volume Aggregation}
For each weighted sample $q$, we let $\mathcal{B}(q) \subset \mathbb{R}^3$ denote the volume occupied by the robot's kinematic links. We project this volume onto the grid and accumulate the weight:
\begin{equation}
\mathcal{H}(v) \leftarrow \mathcal{H}(v) + w(q), \quad \forall v \in \text{Voxelize}(\mathcal{B}(q)) \cap \mathcal{V}_{\mathrm{free}}.
\end{equation}
The resulting $\mathcal{H}$ encodes the density of robust trajectories within the workspace.

\subsection{Online: Adversarial Occupancy Maximization}
\label{sec:method:online}

Given $q_{\mathrm{init}}$ and $\mathcal{G}$, the adversary seeks to instantiate an obstacle set $\mathcal{O}(\theta)$ that maximally covers the heatmap mass relevant to the current task.

\subsubsection{Task-Relevant Candidate Selection}
To satisfy {placement feasibility} (Constraint C3), we define a candidate set $\mathcal{X}_{\text{cand}}$ restricted to a task-aligned corridor connecting $x_{\mathrm{start}}$ and $\mathcal{G}$. Voxels within a safety margin $d_{\text{safe}}$ of $q_{\mathrm{init}}$ or $\mathcal{G}$ are pruned to avoid trivial infeasibility.

\subsubsection{Budgeted Maximum-Coverage Optimization}
We model obstacle placement as a submodular maximization problem~\cite{khuller1999MaxCover}. Let $\mathcal{U}(c)$ be the voxels occupied by a primitive centered at $c$. The adversary solves:
\begin{equation}
\begin{aligned}
    \max_{\mathcal{C} \subseteq \mathcal{X}_{\text{cand}}} \quad & \sum_{v \in \bigcup_{c \in \mathcal{C}} \mathcal{U}(c)} \mathcal{H}(v) \\
    \text{s.t.} \quad & |\mathcal{C}| \le M, \quad \forall c_i, c_j \in \mathcal{C}: \text{dist}(c_i, c_j) \ge d_{\min}.
\end{aligned}
\label{eq:max_cover}
\end{equation}
The constraint $d_{\min}$ ensures {spatial diversity} by preventing obstacles from overlapping excessively. This forces the generated constraints to obstruct multiple potential trajectories rather than a single path.

\subsubsection{Greedy Attack Generation}
The optimization in Eq.~\eqref{eq:max_cover} is NP-hard; however, since the objective is submodular, a greedy strategy yields a $(1-1/e)$ approximation. The iterative selection process is summarized in {Algorithm~\ref{alg:online}}. By selecting the center $c^*$ that maximizes the marginal heatmap coverage in each step, the adversary efficiently constructs a global constraint field that occludes the robot's most robust motion options.

\begin{algorithm}[t]
\caption{Adversarial Occupancy Maximization}
\label{alg:online}
\begin{algorithmic}[1]
\Require Heatmap $\mathcal{H}$; States $q_{\mathrm{init}}, \mathcal{G}$; Budget $M$; Primitive size $s$
\Ensure Adversarial Obstacles $\mathcal{O}$
\State $\mathcal{S} \leftarrow \text{DefineSearchRegion}(q_{\mathrm{init}}, \mathcal{G}, d_{\text{safe}})$ \Comment{Constraint C3}
\State $\mathcal{X} \leftarrow \{v \in \mathcal{V}_{\mathrm{free}} \mid \mathcal{H}(v) > 0 \} \cap \mathcal{S}$
\State $\mathcal{O} \leftarrow \emptyset, \quad \mathcal{V}_{\text{covered}} \leftarrow \emptyset$
\For{$m = 1$ to $M$}
    \If{$\mathcal{X} = \emptyset$} \textbf{break} \EndIf
    \State $c^* \leftarrow \arg\max_{c \in \mathcal{X}} \sum_{v \in \mathcal{U}(c) \setminus \mathcal{V}_{\text{covered}}} \mathcal{H}(v)$
    \State $\mathcal{O} \leftarrow \mathcal{O} \cup \{\text{Primitive}(c^*, s)\}$
    \State $\mathcal{V}_{\text{covered}} \leftarrow \mathcal{V}_{\text{covered}} \cup \mathcal{U}(c^*)$
    \State $\mathcal{X} \leftarrow \{c \in \mathcal{X} \mid \|c - c^*\| \ge d_{\min} \}$ \Comment{Spatial diversity}
\EndFor
\State \Return $\mathcal{O}$
\end{algorithmic}
\end{algorithm}

\section{Experimental Results}
\label{sec:experiments}

We evaluate the proposed planner-agnostic adversarial framework in both high-fidelity simulation and real-robot deployments. Our experiments are designed to answer:
\textbf{(RQ1)} How effective is the attack against classical  planners?
\textbf{(RQ2)} What obstacle budget is required to induce specific failure rates?
\textbf{(RQ3)} Can kinematics-derived obstacles transfer to vision-language-action (VLA) policies?
\textbf{(RQ4)} How does our efficiency compare to planner-in-the-loop baselines?
\textbf{(RQ5)} Can the framework transfer to physical robots under real sensing and modeling noise?

\subsection{Experimental Setup}

\paragraph{Robotic Platform and Simulation}
We use a 7-DoF Franka Emika Panda as the target platform. Simulations are conducted in NVIDIA Isaac Sim~\cite{makoviychuk2isaac}, which provides high-fidelity physics and collision checking. The planning stack is built on ROS~2~\cite{macenski2022ros2} and MoveIt~2~\cite{coleman2014moveit} with OMPL~\cite{sucan2012ompl} planners. Forward kinematics and Jacobians are parsed and evaluated via the Pinocchio library.

\begin{figure*}[!t]
    \centering
    \includegraphics[width=0.95\linewidth]{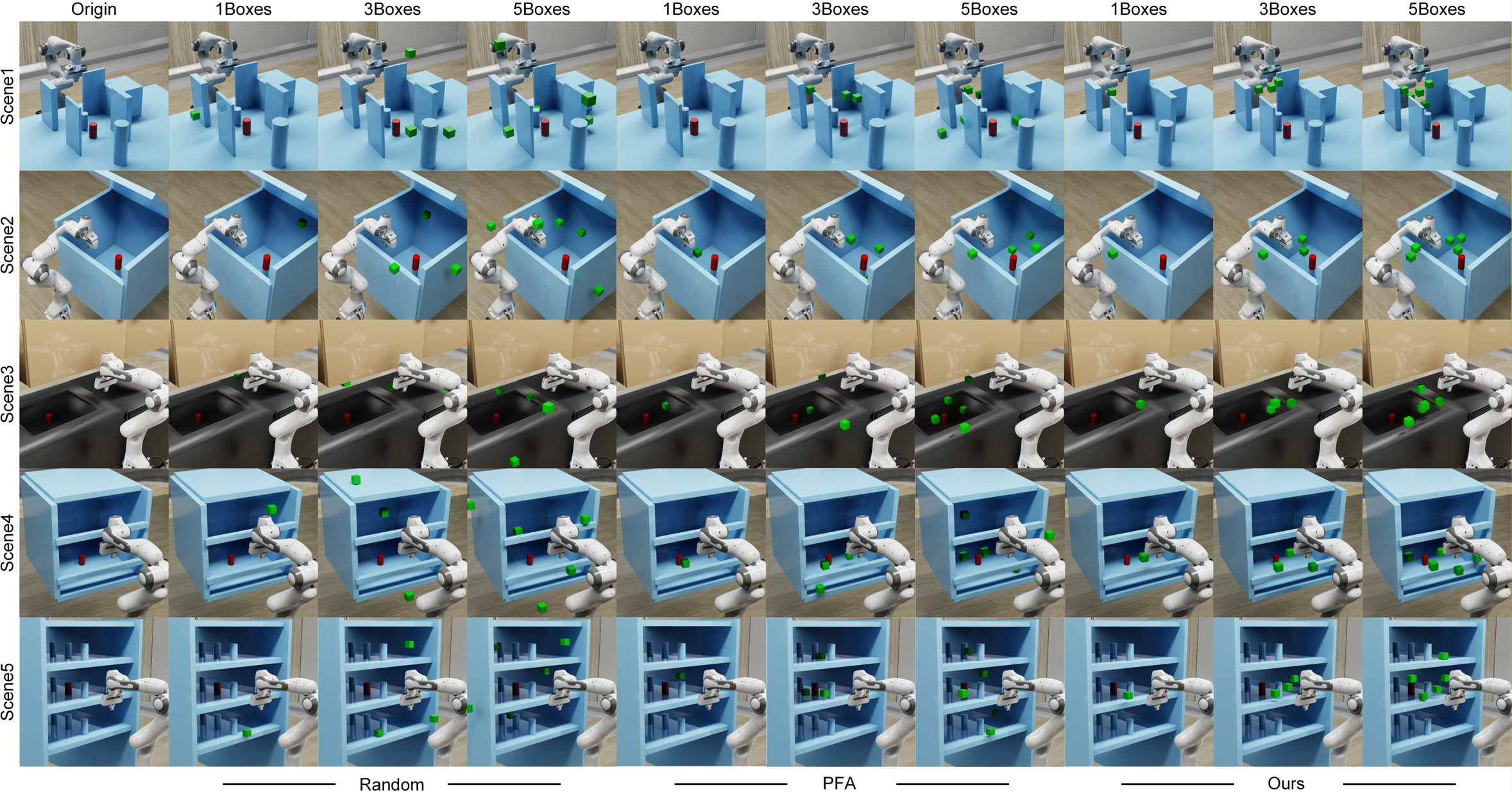}
    \vspace{-3mm}
    \caption{
    Qualitative visualization of adversarial obstacle placement in five tabletop scenes under budgets of 1, 3, and 5.
    }
\label{fig:vis_adv}
\end{figure*}

\paragraph{Scene Design}
We generate five tabletop manipulation scenes (Scene~1--5) using MotionBenchMaker~\cite{chamzas2022-motion-bench-maker}, covering varying clutter patterns and goal regions. Due to space constraints, we report detailed quantitative benchmarks on three representative scenes (Scene~1--3), while comprehensive qualitative evaluations across all five scenes are provided in Fig.~\ref{fig:vis_adv}. 


\paragraph{Planners and Protocol}
To demonstrate planner agnosticism, we evaluate three widely used sampling-based planners: RRT~\cite{kuffner2000rrtconnect}, PRM$^\ast$~\cite{karaman2011sampling}, and BKPIECE~\cite{sucan2012kpiece}. Each planner is given a strict time budget of $5.0$\,s with a single attempt per query. For each (scene, planner, method, budget) configuration, we report averages over $500$ independent runs.

\paragraph{Implementation Details}
We construct the kinematic occupancy heatmap with a voxel resolution of $\delta=0.02$\,m. Adversarial obstacles are modeled as axis-aligned box primitives with a fixed edge length of $s=0.05$\,m and a constant orientation. For the offline heatmap construction, we sample $N_{\text{off}}=15{,}000$ collision-free joint configurations. The discrete workspace volume $\mathcal{V}$ is defined by the bounding box of the robot's swept volume across these valid samples, augmented with a padding margin to fully encompass task-relevant regions. For the heuristic weights in Eq.~(3), we set the manipulability saturation parameter to $\kappa=0.01$ and use equal weighting by setting $\gamma_1 = \gamma_2 = 1$. During the online candidate generation, we prune voxels within a safety margin of $d_{\text{safe}}=0.1$\,m from both $q_{\text{init}}$ and the goal region $\mathcal{G}$. Additionally, candidate centers are restricted to a task-aligned corridor of radius $r_{\text{corr}}=0.3$\,m connecting the start position $x_{\text{start}}$ to $\mathcal{G}$. To ensure spatial diversity among the generated constraints, we enforce a minimum separation of $d_{\min}=0.05$\,m between obstacle centers. We evaluate our framework under budgets of $1$, $3$, and $5$ obstacles.

\paragraph{Baselines}
We consider three baselines.
\begin{itemize}\setlength\itemsep{0.15em}
    \item \textbf{Random}: obstacles uniformly sampled in the valid search region.
    \item \textbf{PFA}: the planner-in-the-loop physical attack of Wu \emph{et al.}~\cite{wu2024characterizing}, which iteratively queries a target planner to optimize obstacle placement.
    \item \textbf{PFA++}: our tolerance-aware adaptation of PFA for fair comparison. Since PFA internally uses an RRT-based evaluation, PFA and PFA++ are mathematically equivalent when evaluated on RRT.
\end{itemize}



\paragraph{Evaluation Metrics} 

We report five metrics capturing attack success and the quality of the surviving plans:
\begin{itemize}\setlength\itemsep{0.15em}
    \item \textbf{Planning Success Rate (PSR)}: Fraction of successful plans out of $500$ trials; lower values indicate a more effective attack.
    \item \textbf{Singularity Margin (SMR, $\times 10^{-2}$)}: Minimum Yoshikawa manipulability $m(q)=\sqrt{\det(J(q)J(q)^\top)}$ along a successful trajectory; lower values indicate the planner is driven closer to kinematic singularities.
    \item \textbf{Clearance Margin (CMR, $\times 10^{-2}$)}: Estimated lower bound of joint-space distance to collision via randomized bi-directional search; lower values indicate narrower free-space corridors.
    \item \textbf{Joint Deviation Score (JDS)}: Average joint-space deviation from the initial configuration along the trajectory; higher values indicate larger detours forced by the adversarial obstacles.
    \item \textbf{Planning Time (PT)}: Total computation time required to return a planning result; failures are recorded as the maximum budget ($5.0$\,second).
\end{itemize}

\subsection{Attack Performance Against Classical Planners (RQ1)}

\begin{table*}[!t]
\centering
\scriptsize
\setlength{\tabcolsep}{3.9pt}
\renewcommand{\arraystretch}{0.95}
\caption{Quantitative evaluation of attack performance across different scenes, planners, and obstacle budgets. SMR and CMR values are scaled by $10^{-2}$. ``--'' denotes non-applicable configurations; ``N/A'' indicates metrics undefined due to complete failure ($\mathrm{PSR}=0$). 
Best values are in bold.
}
\vspace{-3mm}
\label{tab:planner_attack_main}
\begin{tabular}{c c c ccccc ccccc ccccc}
\toprule
\multirow{2}{*}{Scene} & \multirow{2}{*}{Planner} & \multirow{2}{*}{Method} &
\multicolumn{5}{c}{1 Obstacle} & \multicolumn{5}{c}{3 Obstacles} & \multicolumn{5}{c}{5 Obstacles} \\
\cmidrule(lr){4-8}\cmidrule(lr){9-13}\cmidrule(lr){14-18}
& & &
PSR & SMR & CMR & JDS & PT &
PSR & SMR & CMR & JDS & PT &
PSR & SMR & CMR & JDS & PT \\
\midrule

\multirow{12}{*}{Scene~1} & \multirow{4}{*}{RTT} & Random & 88.00 & 5.008 & 3.191 & 2.094 & 0.659 & 87.00 & 4.906 & 3.083 & 2.143 & 0.518 & 85.10 & 5.255 & 2.638 & 2.105 & 0.520 \\
 &  & PFA & 89.20 & 5.081 & 3.261 & 2.779 & 0.522 & 76.40 & 5.163 & 3.261 & 2.042 & 1.145 & 53.00 & 5.217 & 1.680 & 1.958 & 2.355 \\
 &  & PFA++ & -- & -- & -- & -- & -- & -- & -- & -- & -- & -- & -- & -- & -- & -- & -- \\
 &  & Ours & \textbf{79.60} & \textbf{4.850} & \textbf{3.139} & \textbf{2.806} & \textbf{0.681} & \textbf{31.80} & \textbf{4.105} & \textbf{0.777} & \textbf{2.263} & \textbf{3.162} & \textbf{0} & N/A & N/A & N/A & \textbf{4.992} \\
\cmidrule(lr){2-18}
 & \multirow{4}{*}{PRM$^\ast$} & Random & 81.00 & 6.281 & 3.571 & 1.624 & 4.980 & 79.40 & 6.249 & 3.226 & 1.609 & 4.970 & 82.20 & 6.384 & 3.520 & 1.620 & 4.970 \\
 &  & PFA & 80.20 & 6.765 & 3.286 & 1.582 & \textbf{4.990} & 82.20 & 6.348 & 3.275 & \textbf{1.582} & 4.990 & 75.20 & 6.054 & 1.865 & 1.589 & 4.940 \\
 &  & PFA++ & \textbf{78.60} & 6.544 & 3.289 & 1.619 & 4.980 & 79.80 & 6.244 & 3.143 & 1.661 & 4.950 & 71.20 & 5.876 & 1.820 & 1.592 & 5.000 \\
 &  & Ours & 80.60 & \textbf{6.194} & \textbf{3.274} & \textbf{1.645} & \textbf{4.990} & \textbf{38.60} & \textbf{4.346} & \textbf{0.871} & 2.291 & \textbf{5.000} & \textbf{0} & N/A & N/A & N/A & \textbf{5.000} \\
\cmidrule(lr){2-18}
 & \multirow{4}{*}{BKPIECE} & Random & 77.00 & 5.519 & 3.006 & \textbf{2.068} & 0.500 & 77.20 & 5.112 & 3.068 & 2.015 & 0.484 & 79.40 & 5.463 & 2.975 & 2.062 & 0.523 \\
 &  & PFA & 72.40 & 5.341 & 3.441 & 2.005 & 0.756 & 67.80 & 5.262 & 2.861 & 1.995 & 1.235 & 52.80 & 4.837 & 1.424 & 1.979 & 2.146 \\
 &  & PFA++ & 71.60 & \textbf{5.203} & 3.389 & 2.029 & 0.771 & 69.40 & 5.251 & 2.844 & 2.026 & 1.242 & 51.00 & 4.664 & 1.399 & 2.190 & 2.208 \\
 &  & Ours & \textbf{66.40} & 5.476 & \textbf{2.688} & 2.090 & \textbf{0.795} & \textbf{22.40} & \textbf{4.154} & \textbf{0.727} & \textbf{2.277} & \textbf{3.326} & \textbf{0} & N/A & N/A & N/A & \textbf{5.000} \\
\midrule

\multirow{12}{*}{Scene~2} & \multirow{4}{*}{RTT} & Random & 84.80 & 4.843 & 3.289 & 2.245 & 0.540 & 84.20 & 4.667 & 3.613 & 2.225 & 0.509 & 84.40 & 4.814 & 3.280 & 2.229 & 0.554 \\
 &  & PFA & 78.20 & \textbf{4.605} & \textbf{2.317} & \textbf{2.319} & 0.565 & 69.20 & \textbf{4.380} & 1.343 & 2.405 & 1.684 & 48.60 & 3.949 & 1.090 & 2.641 & 1.424 \\
 &  & PFA++ & -- & -- & -- & -- & -- & -- & -- & -- & -- & -- & -- & -- & -- & -- & -- \\
 &  & Ours & \textbf{70.60} & 4.798 & 2.608 & 2.206 & \textbf{0.654} & \textbf{59.20} & 4.740 & \textbf{1.162} & \textbf{2.463} & \textbf{1.739} & \textbf{0} & N/A & N/A & N/A & \textbf{5.000} \\
\cmidrule(lr){2-18}
 & \multirow{4}{*}{PRM$^\ast$} & Random & 78.00 & 5.905 & 3.007 & 1.805 & 4.980 & 79.80 & 6.016 & 3.031 & 1.791 & 4.686 & 74.60 & 6.002 & 2.816 & 1.844 & 5.000 \\
 &  & PFA & 73.80 & 6.134 & 2.680 & 1.912 & 4.990 & 67.20 & 5.834 & 2.042 & 2.041 & 4.972 & 49.00 & 5.258 & 1.654 & 2.365 & 5.000 \\
 &  & PFA++ & \textbf{71.90} & \textbf{5.893} & 2.471 & \textbf{1.962} & \textbf{5.000} & 64.00 & 5.793 & 2.034 & 2.138 & \textbf{5.000} & 47.60 & 5.170 & 1.598 & 2.471 & 5.000 \\
 &  & Ours & 74.60 & 6.304 & \textbf{2.285} & 1.802 & \textbf{5.000} & \textbf{60.80} & \textbf{5.718} & \textbf{1.943} & \textbf{2.199} & \textbf{5.000} & \textbf{0} & N/A & N/A & N/A & \textbf{5.000} \\
\cmidrule(lr){2-18}
 & \multirow{4}{*}{BKPIECE} & Random & 75.40 & 4.165 & 2.256 & 2.293 & 0.548 & 75.00 & 4.379 & 2.221 & 2.267 & 0.546 & 73.20 & 4.290 & 2.142 & 2.329 & 0.549 \\
 &  & PFA & 66.00 & 4.152 & 1.835 & 2.382 & 0.537 & 53.40 & \textbf{4.104} & 1.299 & 2.394 & 1.797 & 39.80 & 3.663 & 1.030 & 2.590 & 1.720 \\
 &  & PFA++ & 63.20 & 4.062 & 1.772 & \textbf{2.417} & 0.620 & 51.00 & 4.194 & 1.196 & 2.400 & 1.902 & 37.40 & 3.599 & 1.019 & 2.370 & 1.919 \\
 &  & Ours & \textbf{61.80} & \textbf{4.038} & \textbf{1.756} & 2.242 & \textbf{0.645} & \textbf{37.00} & 4.067 & \textbf{1.144} & \textbf{2.429} & \textbf{2.108} & \textbf{0} & N/A & N/A & N/A & \textbf{5.000} \\
\midrule

\multirow{12}{*}{Scene~3} & \multirow{4}{*}{RTT} & Random & 95.40 & 5.051 & 4.100 & 1.852 & 0.451 & 94.40 & 4.981 & 3.973 & 1.786 & 0.502 & 94.40 & 4.667 & 3.544 & 1.891 & 0.527 \\
 &  & PFA & 92.40 & 4.959 & 4.200 & 1.845 & 0.472 & 78.80 & 4.919 & 1.996 & 1.921 & 1.297 & 54.20 & 4.346 & 2.459 & 1.930 & 2.504 \\
 &  & PFA++ & -- & -- & -- & -- & -- & -- & -- & -- & -- & -- & -- & -- & -- & -- & -- \\
 &  & Ours & \textbf{83.20} & \textbf{4.265} & \textbf{1.960} & \textbf{2.031} & \textbf{0.503} & \textbf{65.00} & \textbf{4.132} & \textbf{0.901} & \textbf{2.269} & \textbf{1.724} & \textbf{0} & N/A & N/A & N/A & \textbf{4.999} \\
\cmidrule(lr){2-18}
 & \multirow{4}{*}{PRM$^\ast$} & Random & \textbf{98.40} & 6.899 & 4.513 & 1.065 & 4.874 & 98.80 & 6.780 & 4.593 & 1.066 & 4.841 & 99.00 & 6.742 & 4.486 & 1.070 & 4.894 \\
 &  & PFA & 98.80 & 6.768 & 4.732 & 1.062 & 4.799 & 86.20 & 6.849 & 1.916 & 1.222 & 4.871 & 59.60 & 5.635 & 2.580 & 1.142 & 4.990 \\
 &  & PFA++ & 95.20 & 6.506 & 4.546 & 1.177 & 4.836 & 84.40 & 6.709 & 1.860 & 1.404 & \textbf{4.941} & 57.40 & 5.450 & 2.600 & 1.203 & 5.000 \\
 &  & Ours & \textbf{71.20} & \textbf{6.113} & \textbf{1.832} & \textbf{1.319} & \textbf{4.972} & \textbf{47.40} & \textbf{5.246} & \textbf{0.878} & \textbf{1.848} & 4.836 & \textbf{0} & N/A & N/A & N/A & \textbf{5.000} \\
\cmidrule(lr){2-18}
 & \multirow{4}{*}{BKPIECE} & Random & 95.80 & 5.257 & 4.111 & 1.820 & 0.513 & 92.00 & 5.138 & 3.814 & 1.785 & 0.539 & 93.40 & 5.215 & 3.613 & 1.780 & 0.541 \\
 &  & PFA & 94.00 & 5.248 & 3.873 & 1.779 & 0.462 & 69.60 & 4.963 & 1.747 & 1.885 & 1.478 & 41.60 & 4.882 & 2.438 & 2.081 & 2.531 \\
 &  & PFA++ & 91.80 & 5.086 & 3.667 & 1.808 & 0.501 & 64.40 & 4.653 & 1.792 & 1.842 & 1.505 & 39.50 & 4.887 & 2.338 & 1.825 & 2.781 \\
 &  & Ours & \textbf{80.60} & \textbf{4.082} & \textbf{1.808} & \textbf{2.132} & \textbf{0.547} & \textbf{55.20} & \textbf{4.353} & \textbf{0.942} & \textbf{2.178} & \textbf{2.068} & \textbf{0} & N/A & N/A & N/A & \textbf{4.742} \\
\bottomrule
\end{tabular}%
\vspace{-1.5mm}
\end{table*}

\paragraph{Overall Trends}
Table~\ref{tab:planner_attack_main} shows that our method consistently achieves the lowest PSR across almost all scene--planner combinations, and its advantage becomes dominant as the obstacle budget increases. This behavior aligns with our design: rather than optimizing against a specific planner’s sampling pattern, we directly target high-occupancy swept volumes that are shared across feasible configurations. As a result, obstacles placed by our method tend to obstruct the task-level feasible manifold more globally.

\paragraph{Performance Across Obstacle Budgets}
At a low budget (1 obstacle), absolute gaps between methods can be small in easy settings, but our framework already demonstrates consistent improvements in difficult cases. For instance, on PRM$^\ast$ in Scene~3, Random and PFA maintain a PSR close to 1, indicating that a single randomly placed or planner-coupled obstacle rarely hits a true bottleneck. In contrast, our method reduces the PSR to $0.712$, proving that the heatmap can identify structurally important passages even with a single insertion. As the budget increases to 3 obstacles, this advantage is significantly amplified. In Scene~1 under BKPIECE, our PSR drops to $0.224$ while the best baseline remains above $0.67$. Finally, at a high budget (5 obstacles), our method achieves complete occlusion ($\mathrm{PSR}=0$) across all nine scene-planner configurations. In contrast, planner-in-the-loop baselines still leave notable residual success. This performance gap highlights that obstacles derived from kinematic occupancy can effectively cover global workspace bottlenecks, whereas iterative planner-coupled optimization often overfits to specific sampling patterns and misses alternative feasible corridors.

\paragraph{Progressive Degradation of Plan Quality}
Even before reaching $\mathrm{PSR}=0$, the auxiliary metrics demonstrate a consistent pattern of trajectory degradation: (i) \textbf{CMR decreases}, indicating that the remaining feasible motions are squeezed into precarious, narrow passages; (ii) \textbf{SMR decreases}, showing that the planner is driven toward less dexterous, near-singular postures; (iii) \textbf{JDS increases}, implying longer detours; and (iv) \textbf{PT increases}, reflecting substantially harder search scenarios. For example, under BKPIECE in Scene~3, increasing our budget from 1 to 3 reduces CMR from $1.808$ to $0.942$ while PT sharply increases, revealing that even when the victim planner succeeds, it does so under severely compromised safety and kinematic margins.

\subsection{Attack Efficiency on Classical Planners (RQ2)}

\begin{table}[!t]
\centering
\scriptsize
\setlength{\tabcolsep}{4.0pt}
\renewcommand{\arraystretch}{0.95}
\caption{Comparative analysis of attack efficiency. Minimum obstacle budget required to reduce the PSR to specified thresholds.}
\vspace{-3mm}
\label{tab:success_rate_thresholds}
\begin{tabular}{c c c ccccc}
\toprule
\multirow{2}{*}{Scene} & \multirow{2}{*}{Planner} & \multirow{2}{*}{Method} & \multicolumn{5}{c}{Success Rate} \\
\cmidrule(lr){4-8}
& & & 80\% & 60\% & 40\% & 20\% & 0\% \\
\midrule

\multirow{12}{*}{Scene~1}
& \multirow{4}{*}{RTT}
& Random  & 6.2 & 15.7 & 25.6 & 35.3 & 50.8 \\
& & PFA    & 2.5 & 4.5  & 6.5  & 9.0  & 12.4 \\
& & PFA++  & --  & --   & --   & --   & --   \\
& & Ours    & \textbf{1.0} & \textbf{1.9}  & \textbf{2.6}  & \textbf{3.6}  & \textbf{4.1}  \\
\cmidrule(lr){2-8}

& \multirow{4}{*}{PRM$^\ast$}
& Random  & 2.5 & 12.4 & 20.2 & 38.4 & 55.6 \\
& & PFA    & 1.5 & 8.6  & 10.2 & 14.1 & 18.4 \\
& & PFA++  & 1.2 & 6.3  & 9.6  & 13.7 & 16.6 \\
& & Ours    & \textbf{1.1} & \textbf{1.8}  & \textbf{2.9}  & \textbf{3.8}  & \textbf{4.7}  \\
\cmidrule(lr){2-8}

& \multirow{4}{*}{BKPIECE}
& Random  & \textbf{1.0} & 12.2 & 19.5 & 27.4 & 45.8 \\
& & PFA    & \textbf{1.0} & 4.5  & 6.3  & 9.9  & 12.3 \\
& & PFA++  & \textbf{1.0} & 4.8  & 6.8  & 9.5  & 13.2 \\
& & Ours    & \textbf{1.0} & \textbf{1.3}  & \textbf{2.2}  & \textbf{3.2}  & \textbf{4.7}  \\
\midrule

\multirow{12}{*}{Scene~2}
& \multirow{4}{*}{RTT}
& Random  & 8.0 & 19.7 & 23.9 & 31.3 & 48.4 \\
& & PFA    & \textbf{1.0} & 3.7  & 5.9  & 8.5  & 12.2 \\
& & PFA++  & --  & --   & --   & --   & --   \\
& & Ours    & \textbf{1.0} & \textbf{2.8}  & \textbf{3.6}  & \textbf{4.3}  & \textbf{4.8}  \\
\cmidrule(lr){2-8}

& \multirow{4}{*}{PRM$^\ast$}
& Random  & \textbf{1.0} & 20.7 & 25.5 & 35.0 & 51.2 \\
& & PFA    & \textbf{1.0} & 3.8  & 6.0  & 8.4  & 11.6 \\
& & PFA++  & \textbf{1.0} & 4.3  & 5.8  & 8.0  & 10.7 \\
& & Ours    & \textbf{1.0} & \textbf{3.1}  & \textbf{3.7}  & \textbf{4.4}  & \textbf{4.9}  \\
\cmidrule(lr){2-8}

& \multirow{4}{*}{BKPIECE}
& Random  & \textbf{1.0} & 13.0 & 22.8 & 30.4 & 39.4 \\
& & PFA    & \textbf{1.0} & 2.0  & 4.9  & 7.8  & 10.3 \\
& & PFA++  & \textbf{1.0} & 1.8  & 4.5  & 7.1  & 9.5  \\
& & Ours    & \textbf{1.0} & \textbf{1.1}  & \textbf{2.8}  & \textbf{3.9}  & \textbf{4.5}  \\
\midrule

\multirow{12}{*}{Scene~3}
& \multirow{4}{*}{RTT}
& Random  & 14.9 & 32.1 & 42.7 & 69.7 & 91.1 \\
& & PFA    & 2.8  & 4.4  & 6.3  & 8.7  & 11.9 \\
& & PFA++  & --   & --   & --   & --   & --   \\
& & Ours    & \textbf{1.3}  & \textbf{3.4}  & \textbf{4.1}  & \textbf{4.6}  & \textbf{4.9}  \\
\cmidrule(lr){2-8}

& \multirow{4}{*}{PRM$^\ast$}
& Random  & 21.0 & 39.7 & 54.8 & 71.7 & 93.2 \\
& & PFA    & 3.6  & 4.9  & 6.7  & 8.5  & 10.9 \\
& & PFA++  & 3.4  & 4.8  & 6.4  & 8.0  & 10.0 \\
& & Ours    & \textbf{1.0}  & \textbf{1.9}  & \textbf{3.3}  & \textbf{4.2}  & \textbf{4.4}  \\
\cmidrule(lr){2-8}

& \multirow{4}{*}{BKPIECE}
& Random  & 10.5 & 21.3 & 40.9 & 54.0 & 78.0 \\
& & PFA    & 2.2  & 3.6  & 5.2  & 7.2  & 9.3  \\
& & PFA++  & 1.9  & 3.3  & 4.9  & 6.8  & 8.5  \\
& & Ours    & \textbf{1.0}  & \textbf{2.6}  & \textbf{3.5}  & \textbf{4.3}  & \textbf{4.2}  \\
\bottomrule
\end{tabular}%
\end{table}

\paragraph{Interpretation of Budget Thresholds}
Table~\ref{tab:success_rate_thresholds} provides a direct ``budget-to-failure'' perspective, which is often more informative for safety evaluations than reporting PSR under a fixed budget. Across all scenes and planners, our method reaches each failure threshold using substantially fewer obstacles. For example, to completely block RRT in Scene~1 ($\mathrm{PSR}=0\%$), Random requires $50.8$ obstacles and PFA requires $12.4$, whereas our method requires only $4.1$. In the most challenging configuration (Scene~3 under PRM$^\ast$), complete occlusion demands $93.2$ obstacles for Random, $10.9$ for PFA, and $10.0$ for PFA++, compared to a mere $4.4$ for our method. Crucially, our approach never requires more than $5$ obstacles to achieve absolute planning failure across any tested configuration. This validates our core premise: a small number of optimally placed constraints over the kinematic occupancy field is sufficient to completely paralyze the manipulator.

\subsection{Transferability to VLA-based Policies (RQ3)}

\begin{table*}[!t]
\centering
\scriptsize
\setlength{\tabcolsep}{3.4pt}
\caption{Quantitative comparison of attack transferability to VLA policies (OpenVLA and $\pi_{0.5}$). }
\vspace{-3mm}
\label{tab:openvla_pi_obstacles}
\begin{tabular}{c c cccc cccc cccc}
\toprule
\multirow{2}{*}{Model} & \multirow{2}{*}{Method} &
\multicolumn{4}{c}{1 Obstacle} &
\multicolumn{4}{c}{3 Obstacles} &
\multicolumn{4}{c}{5 Obstacles} \\
\cmidrule(lr){3-6}\cmidrule(lr){7-10}\cmidrule(lr){11-14}
& &
PSR & SMR & CMR & JDS &
PSR & SMR & CMR & JDS &
PSR & SMR & CMR & JDS \\
\midrule

\multirow{4}{*}{OpenVLA}
& Random  & 45.60 & 5.468 & 0.951 & 1.832 & 43.60 & 5.530 & 0.825 & 1.901 & 43.20 & 5.408 & 0.851 & 2.002 \\
& PFA(RRT)    & 36.40 & 5.312 & 0.848 & 2.057 & 18.80 & 5.309 & 0.790 & 2.152 & \textbf{0}     & N/A   & N/A   & N/A   \\
& PFA++(PRM*)  & 35.80    & \textbf{5.122}    & 0.813    & 2.029    & 17.90    & 5.226    & 0.771    & 2.190    & \textbf{0}    & N/A    & N/A    & N/A    \\
& Ours    & \textbf{34.00} & 5.335 & \textbf{0.792} & \textbf{2.003} & \textbf{0}     & N/A   & N/A   & N/A   & \textbf{0}     & N/A   & N/A   & N/A   \\
\midrule

\multirow{4}{*}{Pi 0.5}
& Random  & 58.60 & 5.612 & 1.310 & 1.988 & 58.40 & 5.456 & 1.262 & 2.071 & 52.20 & 5.690 & 0.923 & 2.196 \\
& PFA(RRT)    & 51.20 & 5.679 & 1.103 & 1.857 & 42.60 & 5.317 & 0.933 & 2.140 & 0     & N/A   & N/A   & N/A   \\
& PFA++(PRM*)  & 50.20 & 5.601 & 1.154 & 1.973 & 39.00 & 5.286 & 0.905 & 2.251 & 0     & N/A   & N/A   & N/A   \\
& Ours    & \textbf{47.40} & \textbf{5.323} & \textbf{1.047} & \textbf{2.034} & \textbf{14.20} & \textbf{5.101} & \textbf{0.877} & \textbf{2.306} & \textbf{0}     & N/A   & N/A   & N/A   \\
\bottomrule
\end{tabular}%
\end{table*}

\paragraph{Transfer Results and Implications}
Table~\ref{tab:openvla_pi_obstacles} reports the attack performance on two representative VLA policies, OpenVLA~\cite{kim24openvla} and $\pi_{0.5}$~\cite{black2025pi_}. Our obstacles are generated without access to the victim policy’s network architecture, training data, or gradients, relying solely on robot kinematic occupancy. Despite this strict black-box setting, the attack remains highly effective.

Against OpenVLA, a single obstacle already reduces the PSR to $0.340$ (compared to $0.364$ for PFA). With three obstacles, our method achieves complete occlusion ($\mathrm{PSR}=0$), whereas PFA still leaves a residual success rate of $0.188$. Against $\pi_{0.5}$, our method yields a PSR of $0.142$ under a 3-obstacle budget, substantially outperforming PFA ($0.426$) and PFA++ ($0.390$). At a budget of 5 obstacles, the success rate collapses entirely. These results suggest that kinematic bottlenecks constitute a shared physical vulnerability across classical sampling-based planners and end-to-end learned policies: while their decision-making mechanisms differ, both ultimately operate within the exact same geometric and kinematic feasibility constraints.

\paragraph{Qualitative Failure Modes of VLA Policies}
We additionally observe qualitatively different failure behaviors compared to classical planners. When the task becomes infeasible, classical planners typically return failure explicitly within the planning budget. In contrast, attacked VLA policies often exhibit progressive, cascading failures: the robot first slows down and hesitates near the obstacle, then drifts into contact, and finally enters oscillatory or deadlocked states, see supplementary video. This highlights that current VLA policies may lack explicit geometric feasibility reasoning when the solution manifold is adversarially occluded.

\subsection{Computational Efficiency (RQ4)}

\begin{table}[!t]
\centering
\footnotesize
\setlength{\tabcolsep}{3.4pt}
\caption{Quantitative comparison of offline and online runtimes (in seconds) across different methods.}
\vspace{-3mm}
\label{tab:runtime_comparison}
\begin{tabular}{lccc}
\toprule
{Method} & {Offline Time} & {Online Time} & {Total Time} \\
\midrule
Random                & 0.10    & --       & 0.10    \\
PFA                  & --      & 1224.75  & 1224.75 \\
PFA++ (PRM*)          & --      & 3978.42  & 3978.42 \\
PFA++ (BKPIECE)      & --      & 1326.67  & 1326.67 \\
Ours                  & 1251.47 & 1.27     & 1252.74 \\
\bottomrule
\end{tabular}%
\vspace{-1.5mm}
\end{table}

\begin{figure}[!t]
    \centering
    \includegraphics[width=0.9\linewidth]{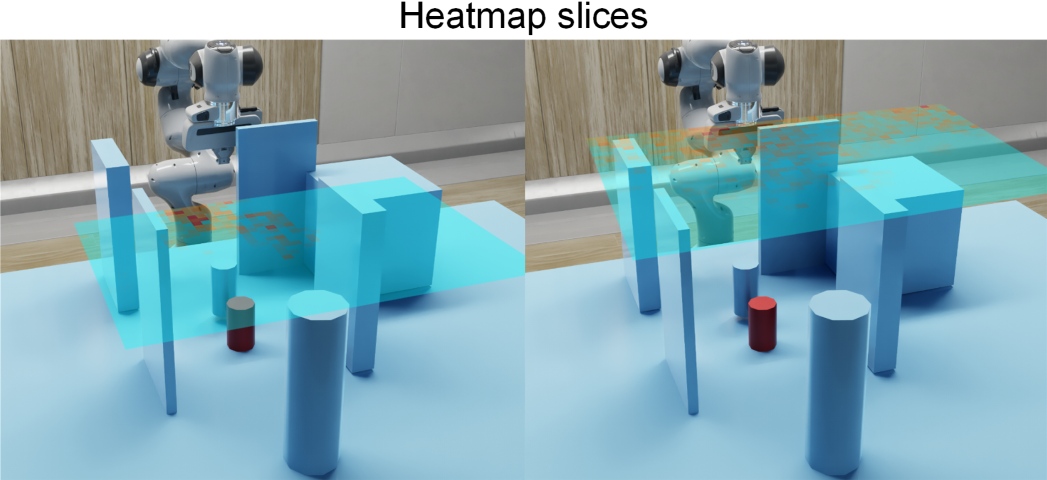}
    \vspace{-3mm}
    \caption{Visualization of the kinematic occupancy heatmap (2D slice). Each voxel stores the swept-volume frequency under collision-free configurations, enabling budgeted maximum-coverage obstacle placement.}
\label{fig:vis_heatmap}
\end{figure}

Table~\ref{tab:runtime_comparison} compares the computational runtimes across different attack methods. Our framework strategically shifts the heavy computational burden to a one-time offline heatmap construction ($\approx 1251$\,s, visualized in Fig.~\ref{fig:vis_heatmap}), enabling extremely fast online attack generation ($\approx 1.27$\,s). This paradigm contrasts sharply with planner-in-the-loop attacks: PFA absorbs its entire computational cost during the online phase, requiring $\approx 1224$\,s of iterative optimization for \emph{every} new task query. PFA++ can be even more exorbitant, requiring $\approx 3978$\,s when coupled with PRM$^\ast$. Crucially, because the offline heatmap is tied to the robot's intrinsic kinematics rather than specific scene layouts, it is constructed once per robotic platform and reused indefinitely across varying tasks and goal regions. This massive amortized advantage makes our approach uniquely suited for near real-time safety evaluations and large-scale robustness benchmarking.

\subsection{Real-World Robotic Deployments (RQ5)}

\begin{figure}[!t]
    \centering
    \includegraphics[width=1\linewidth]{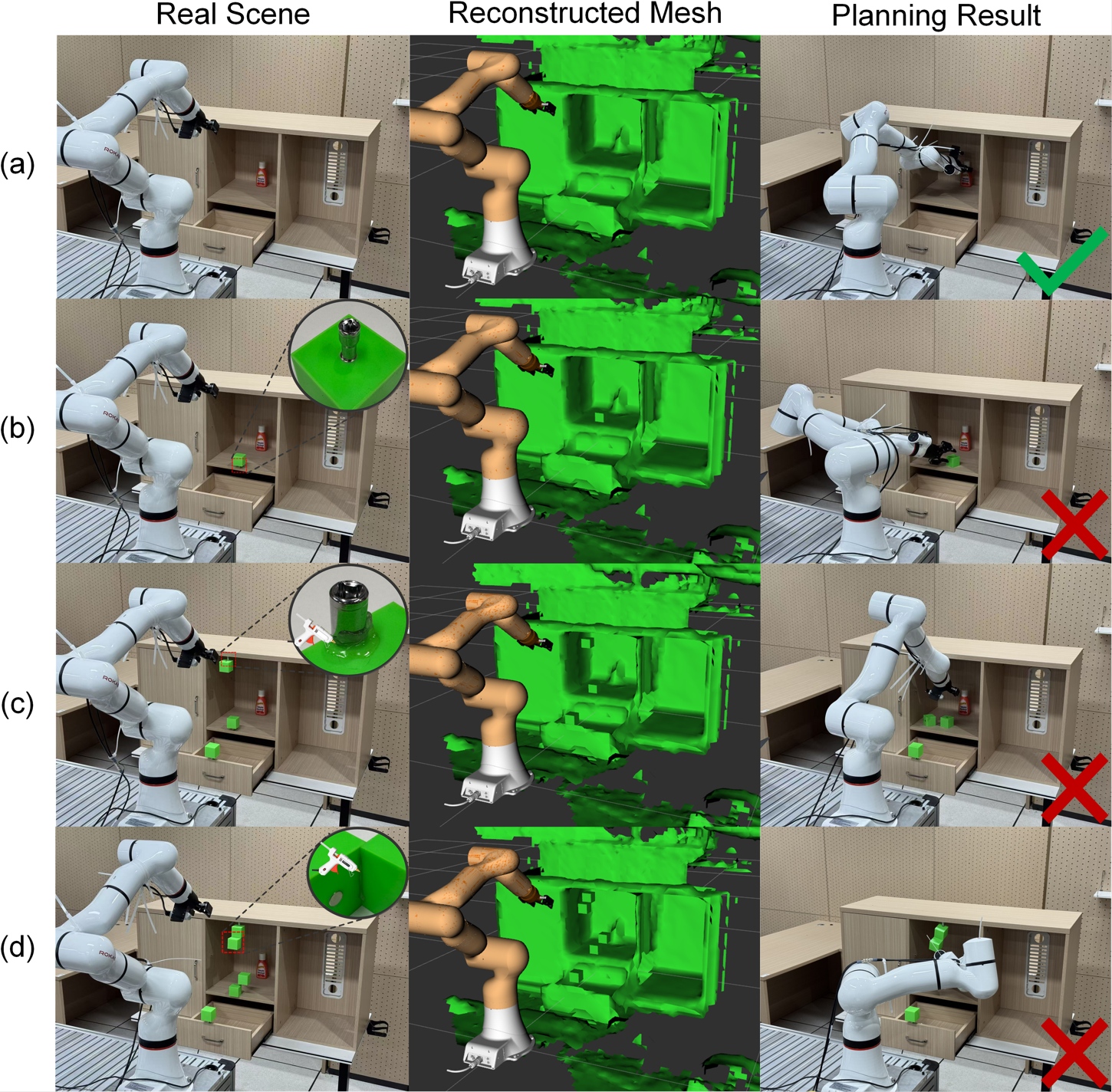}
    \caption{Real-robot deployment on Rokae xMatePro7. (a) Original scene without adversarial obstacles. (b--d) Adversarial scenes with 1/3/5 obstacles.}
\label{fig:real_world}
\end{figure}

To validate the sim-to-real transferability of our framework under realistic sensing constraints, we deployed the attack on a physical Rokae xMatePro7 manipulator (Fig.~\ref{fig:real_world}). The planning scene is constructed directly from real-world RGB-D sensing. Specifically, an Intel RealSense camera captures the workspace, from which we reconstruct a geometric collision mesh (visualized as the green overlay in MoveIt 2). Our adversarial obstacle placement is then executed over this sensed mesh, ensuring the attack is strictly grounded in the robot's actual perception. To physically realize the attack, the generated adversarial obstacles are instantiated using rigidly mounted box primitives, ensuring spatial consistency with the computed geometric constraints while controlling for unwanted motion during repeated trials.

The physical deployments closely mirror our simulation findings. As shown in Fig.~\ref{fig:real_world}(a), without adversarial obstacles, the robot smoothly plans and executes a collision-free trajectory to the goal region. When a single adversarial obstacle is introduced (b), the planner may still succeed but is forced into a narrow, suboptimal path, visibly increasing joint deviations and execution complexity. However, as the obstacle budget increases to three and five (c--d), the planner consistently fails to find a feasible solution within the time budget and safely aborts. This behavior perfectly matches the ``complete occlusion'' phenomenon observed in simulation, confirming that our geometrically derived attack remains highly robust and effective in the real world despite raw sensor noise and execution uncertainties.

\section{Conclusion and Discussion}
\label{sec:conclusion}

In this paper, we have presented a planner-agnostic adversarial framework targeting the  robustness of tolerance-aware manipulation planning. By shifting the focus from rigid goal poses to task-level solution manifolds, our approach utilizes an offline kinematic occupancy heatmap to identify critical workspace regions and an online budgeted maximum-coverage optimization to strategically place obstacles. Unlike prior methods, our framework operates without oracle access to the victim's internal logic, ensuring high efficiency and broad applicability. Extensive experiments in both simulation and real-world scenarios demonstrate that our method effectively occludes the task-level solution space, significantly outperforming planner-in-the-loop baselines in both success rate and computational speed. 

Our results suggest that kinematic bottlenecks form a planner- and policy-agnostic vulnerability surface for tolerance-aware manipulation. Beyond attack generation, the proposed framework can serve as a diagnostic tool to expose where feasibility concentrates and how quickly it collapses under bounded perturbations, providing concrete signals for improving robustness. 
Future work will extend the framework to dynamic environments.









\bibliographystyle{ieeetr}
\bibliography{references}

\end{document}